\newcommand{\ted}{TED }
\newcommand{\ignore}[1]{}

\def\year{2019}\relax
\documentclass[letterpaper]{article} %DO NOT CHANGE THIS
\usepackage{aaai19}  %Required
\usepackage{times}  %Required
\usepackage{helvet}  %Required
\usepackage{courier}  %Required
\usepackage{url}  %Required
\usepackage{graphicx}  %Required
\usepackage{amsmath,amssymb}  %Added by Mike Hind
\usepackage{color}
\usepackage{natbib}
\usepackage{subfig}

\begin{document}
\title{Teaching Meaningful Explanations}
\author{Noel C.\ F.\ Codella,\thanks{These authors contributed equally.}
  Michael Hind,$^*$
  Karthikeyan Natesan Ramamurthy,$^*$  \\
 {\bf \Large Murray Campbell,
  Amit Dhurandhar,
  Kush R.\ Varshney,  
  Dennis Wei,
  Aleksandra Mojsilovi\'c} \\
    IBM Research \\
  Yorktown Heights, NY 10598}
\maketitle
\begin{abstract}
The adoption of machine learning in high-stakes applications such as healthcare and law has lagged in part because predictions are not accompanied by explanations comprehensible to the domain user, who often holds the ultimate responsibility for decisions and outcomes. In this paper, we propose an approach to generate such explanations in which training data is augmented to include, in addition to features and labels, explanations elicited from domain users.  
A joint model is then learned to produce both labels and explanations from the input features.  This simple idea ensures that explanations are tailored to the complexity expectations and domain knowledge of the consumer. Evaluation spans multiple modeling techniques on a game dataset, a (visual) aesthetics dataset, a chemical odor dataset and a Melanoma dataset showing that our approach is generalizable across domains and algorithms. Results demonstrate that meaningful explanations can be reliably taught to machine learning algorithms, and in some cases, also improve modeling accuracy.
\end{abstract}

\noindent 
\section{Introduction}
\label{sec:intro}

New regulations call for automated decision making systems to provide ``meaningful information'' on the logic used to reach conclusions \citep{gdpr-goodman,gdpr-wachter,SelbstP2017}. \citet{SelbstP2017} interpret the concept of ``meaningful information'' as information that should be understandable to the audience (potentially individuals who lack specific expertise), is actionable, and  is flexible enough to support various technical approaches.

For the present discussion, %To facilitate the development and study of explanations, 
we define an explanation as information provided in addition to an output that can be used to verify the output. In the ideal case, an explanation should enable a human user to independently determine whether the output is correct. The requirements of meaningful information have two implications for explanations:

\begin{enumerate}
\item Complexity Match: The {\em complexity of the explanation}
needs to match the complexity capability of the consumer~\citep{too-much,tip}.
For example, an explanation in equation form may be appropriate for a statistician, but not for a nontechnical person~\citep{miller2017inmates}. 

\item Domain Match: An explanation needs to be {\em tailored to the domain}, incorporating the relevant terms of the domain.  For example, an explanation for a medical diagnosis needs to use terms relevant to the physician (or patient) who will be consuming the prediction. 
\end{enumerate}

%Work in social and behavioral sciences \citep{simplicity,miller2017inmates,miller2017explanation-review}
%has found that people prefer explanations that are simpler, more general, and coherent even over more likely ones. 
%Moreover, the European Union GDPR guidelines~\citep{gdpr-interpretation-2017} say: "The controller should find simple ways to tell the data subject about the rationale behind, or the criteria relied on in reaching the decision without necessarily always attempting a complex explanation of the algorithms used or disclosure of the full algorithm."

In this paper, we take this guidance to heart by asking consumers themselves to provide explanations that are \emph{meaningful} to them for their application along with feature/label pairs, where these provided explanations lucidly justify the labels for the specific inputs. We then use this augmented training set to learn models that predict explanations along with labels for new unseen samples.

The proposed paradigm is different from existing methods for local interpretation \citep{montavon2017methods} in that it does not attempt to probe the reasoning process of a \emph{model}. Instead, it seeks to replicate the reasoning process of a human domain user. %The two paradigms can be complementary and used together. %to inform different audiences. The former can tell AI system builders whether the system is behaving as expected and can diagnose problems and suggest improvements. 
The two paradigms share the objective to produce a reasoned explanation, but the  model introspection approach is more well-suited to AI system builders who work with models directly, whereas the teaching explanations paradigm more directly addresses domain users. Indeed, the European Union GDPR guidelines say: ``The controller should find simple ways to tell the data subject about the rationale behind, or the criteria relied on in reaching the decision without necessarily always attempting a complex explanation of the algorithms used or disclosure of the full algorithm.'' More specifically, teaching explanations allows user verification and promotes trust. Verification is facilitated by the fact that the returned explanations are in a form familiar to the user. As predictions and explanations for novel inputs match with a user's intuition, trust in the system will grow accordingly. Under the model introspection approach, while there are certainly cases where model and domain user reasoning match, this does not occur by design and they may diverge in other cases, potentially decreasing trust \citep{Weller2017}.

There are many possible instantiations for this proposed paradigm of teaching explanations. One is to simply expand the label space to be the Cartesian product of the original labels and the elicited explanations. Another approach is to bring together the labels and explanations in a multi-task setting.  The third builds upon the tradition of similarity metrics, case-based reasoning and content-based retrieval.

Existing approaches that only have access to features and labels are unable to find \emph{meaningful} similarities. However, with the advantage of having training features, labels, \emph{and} explanations, we propose to learn feature embeddings 
guided by labels and explanations.  This allows us to infer explanations for new data using nearest neighbor approaches. We present a new objective function to learn an embedding to optimize $k$-nearest neighbor ($k$NN) search for both prediction accuracy as well as holistic human relevancy to enforce that returned neighbors present meaningful information. The proposed embedding approach is easily portable to a diverse set of label and explanation spaces because it only requires a notion of similarity between examples in these spaces. Since any predicted explanation or label is obtained from a simple combination of training examples, \textit{complexity} and \textit{domain} match is achieved with no further effort. We also demonstrate the multi-task instantiation wherein labels and explanations are predicted together from features. In contrast to the embedding approach, we need to change the structure of the ML model
for this method due to the modality and type of the label and explanation space.

We demonstrate the proposed paradigm using the three instantiations on a synthetic tic-tac-toe dataset (See supplement), and publicly-available image aesthetics dataset \citep{kong2016aesthetics}, olfactory pleasantness dataset \citep{olfs}, and melanoma classification dataset \citep{codella2018skin}. Teaching explanations, of course requires a training set that contains explanations.  Since such datasets are not readily available, we use the attributes given with the aesthetics and pleasantness datasets in a unique way: as collections of meaningful explanations. For the melanoma classification dataset, we will use the groupings given by human users described in \citet{codella2018collaborative} as the explanations.

The main contributions of this work are:
\begin{itemize}
\item A new approach for machine learning algorithms to provide \emph{meaningful} explanations that match the complexity and domain of consumers by eliciting training explanations directly from them.  We name this paradigm TED for `Teaching Explanations for Decisions.'
\item Evaluation of several candidate approaches, some of which learn joint embeddings so that the multidimensional topology of a model mimics both the supplied labels and explanations, which are then compared with single-task and multi-task regression/classification approaches.
%\item A methodology to teach explanations by augmenting the label space with explanations.
%\item A candidate kNN based classification method that jointly learns similarities in feature space and explanation space. We adopt a kNN based approach since, it can be readily applied to data in different modalities (tabular, image or text) requiring only a similarity measure with the additional inherent advantage of it being interpretable.
\item Evaluation on disparate datasets with diverse label and explanation spaces demonstrating the efficacy of the paradigm.
\end{itemize}

 \section{Related Work} 
\label{sec-related}

Prior work in providing explanations can be partitioned into several areas:

\begin{enumerate}
\item \label{interp}
Making existing or enhanced models {\em interpretable}, i.e.\ to provide a precise description of how the model determined its
decision (e.g.,~\citet{RibieroSG2016, montavon2017methods, unifiedPI}).

\item \label{2nd-model}
Creating a second, simpler-to-understand model, such as a small number of logical expressions, that mostly 
matches the decisions of the deployed model (e.g.,~\citet{bastani2017interpreting, Caruana:2015}).

\item \label{rationale}
Leveraging ``rationales'', ``explanations'', ``attributes'', or other ``privileged information'' in the training data 
to help improve the accuracy of the
algorithms (e.g.,~\citep{sun-dejong-2005,Zaidan07using-annotator,zaidan-eisner:2008:gen,DBLP:journals/corr/ZhangMW16,McDonnel16why-relevant,rationales,localizedattributes,peng-vision} 
%% Shorter ref list used for AAAI submission
%~\citet{sun-dejong-2005,zaidan-eisner:2008:gen,rationales,peng-vision}). 

\item \label{gen-rationale}
Work in the natural language processing and computer vision domains that generate rationales/explanations derived from input text (e.g., \citet{lei2016rationalizing,Yessenalina:2010:AGA:1858842.1858904,hendricks-2016}).

\item \label{cbir}
Content-based retrieval methods that provide explanations as {\em evidence} employed for a prediction, i.e. $k$-nearest neighbor classification and regression (e.g., \citet{cbir3,cbir4,cbir5,patientsimilarity}).
%nipscbir

\end{enumerate}

The first two groups attempt to precisely describe how a
machine learning decision was made, which is particularly relevant for
AI system builders.  This insight can be used to improve the AI 
system and may serve as the seeds for an explanation to a non-AI expert.
However, work still remains to determine if these seeds are sufficient
to satisfy the needs of a non-AI expert.  In particular, when the underlying features are not human comprehensible, these approaches are inadequate for providing human consumable explanations.

The third group, like this work, leverages additional information (explanations) in the training data, but with different goals.  The third group uses the explanations to create a more accurate model; we leverage the explanations to teach how to generate explanations for new predictions.  

The fourth group seeks to generate textual explanations with predictions. For text classification, this involves selecting the minimal necessary content from a text body that is sufficient to trigger the classification. For computer vision~\citep{hendricks-2016}, this involves utilizing textual captions to automatically generate new textual captions of images that are both descriptive as well as discriminative. While serving to enrich an understanding of the predictions, these systems do not necessarily facilitate an improved ability for a human user to understand system failures.  

The fifth group creates explanations in the form of {\em decision evidence}: using some feature embedding to perform {\em k}-nearest neighbor search, using those {\em k} neighbors to make a prediction, and demonstrating to the user the nearest neighbors and any relevant information regarding them. Although this approach is fairly straightforward and holds a great deal of promise, it has historically suffered from the issue of the semantic gap: distance metrics in the realm of the feature embeddings do not necessarily yield neighbors that are relevant for prediction. More recently, deep feature embeddings, optimized for generating predictions, have made significant advances in reducing the semantic gap. However, there still remains a ``meaning gap'' --- although systems have gotten good at returning neighbors with the same label as a query, they do not necessarily return neighbors that agree with any {\em holistic human measures} of similarity. As a result, users are not necessarily inclined to trust system predictions.

\citet{finale} discuss the societal, moral, and legal expectations of AI explanations, provide
guidelines for the content of an explanation, and recommend that
explanations of AI systems be held to a similar standard as humans.  Our
approach is compatible with their view.   \citet{biran-cotton-2017} provide an excellent overview and taxonomy of explanations and justifications in machine learning.

\citet{miller2017explanation-review} and \citet{miller2017inmates}
argue that explainable AI solutions need to meet the needs of the
users, an area that has been well studied in philosophy, psychology,
and cognitive science.  They provides a brief survey of the most
relevant work in these fields to the area of explainable AI.
They, along with \citet{rsi}, call for more rigor
in this area.

\section{Methods}  \label{methods}

The primary motivation of the TED paradigm is to provide meaningful explanations to consumers by leveraging the consumers' knowledge of what will be meaningful to them.
Section~\ref{sec-problem} formally describes the problem space that defines the TED approach.  One simple learning approach to this problem is to expand the label space to be the Cartesian product of the original labels and the provided explanations.  Although quite simple, this approach has a number of pragmatic advantages in that it is easy to incorporate, it can be used for any learning algorithm, it does not require any changes to the learning algorithm, and does not require owners to make available their algorithm.  It also has the possibility of some indirect benefits because requiring explanations will improve auditability (all decisions will have explanations) and potentially reduce bias in the training set because inconsistencies in explanations may be discovered.

Other instantiations of the TED approach may leverage the explanations to improve model prediction and possibly explanation accuracy.  
Section~\ref{jointpairwise} takes this approach  to learn feature embeddings and explanation embeddings in a joint and aligned way to permit neighbor-based explanation prediction.  It presents a new objective function to learn an embedding to optimize $k$NN search for both prediction accuracy as well as holistic human relevancy to enforce that returned neighbors present meaningful information. We also discuss multi-task learning in the label and explanation space as another instantiation of the TED approach, that we will use for comparisons.
 
\subsection{Problem Description} 
\label{sec-problem}
Let $X\times Y$ denote the input-output space, with $p(x,y)$ denoting the joint distribution over this space, where $(x,y)\in X\times Y$. Then typically, in supervised learning one wants to estimate $p(y|x)$.

In our setting, we have a triple $X\times Y\times E$ that denotes the input space, output space, and explanation space, respectively. We then assume that we have a joint distribution $p(x,y,e)$ over this space, where $(x,y,e)\in X\times Y\times E$. In this setting we want to estimate $p(y,e|x)=p(y|x)p(e|y,x)$. Thus, we not only want to predict the labels $y$, but also the corresponding explanations $e$ for the specific $x$ and $y$ based on historical explanations given by human experts.

The space $E$ in most of these applications is quite different than $X$ and 
has similarities with $Y$ in that it requires human judgment.

We provide methods to solve the above problem. Although these methods can be used even when $X$ is human-understandable, we envision the most impact for applications where this is not the case, such as the olfaction dataset described in Section~\ref{sec-eval}.

\subsection{Candidate Approaches}
\label{jointpairwise}

We propose several candidate implementation approaches to teach labels and explanations from the training data, and predict them for unseen test data. We will describe the baseline regression and embedding approaches. The particular parameters and specific instantiations are provided in Section \ref{sec-eval}.

\subsubsection{Baseline for Predicting $Y$ or $E$}
To set the baseline, we trained a regression (classification) network on the datasets to predict $Y$ from $X$ using the mean-squared error (cross-entropy) loss. This cannot be used to infer $E$ for a novel $X$. A similar learning approach was be used to predict $E$ from $X$. If $E$ is vector-valued, we used multi-task learning.

\subsubsection{Multi-task Learning to Predict $Y$ and $E$ Together}
We trained a multi-task network to predict $Y$ and $E$ together from $X$. Similar to the previous case, we used appropriate loss functions.

\subsubsection{Embeddings to Predict $Y$ and $E$}
We propose to use the activations from the last fully connected hidden layer of the network trained to predict $Y$ or $E$ as embeddings for $X$. Given a novel $X$, we obtain its $k$ nearest neighbors in the embedding space from the training set, and use the corresponding $Y$ and $E$ values to obtain predictions as weighted averages. The weights are determined using a Gaussian kernel on the distances in the embedding space of the novel $X$ to its neighbors in the training set. This procedure is used with all the $k$NN-based prediction approaches.

\subsubsection{Pairwise Loss for Improved Embeddings}
Since our key instantiation is to predict $Y$ and $E$ using the $k$NN approach described above, we propose to improve upon the embeddings of $X$ from the regression network by explicitly ensuring that points with similar $Y$ and $E$ values are mapped close to each other in the embedding space. %our neighborhood is also defined based on $Y$ and $E$. 
For a pair of data points $(a,b)$ with inputs $(x_a,x_b)$, labels $(y_a,y_b)$, and explanations $(e_a,e_b)$, we define the following pairwise loss functions for creating the embedding $f(\cdot)$, where the shorthand for $f(x_i)$ is $f_i$ for clarity below:
\begin{multline}\label{equation-loss-x-y}
L_{x,y}(a,b)\\ = \begin{cases}
1 - \cos(f_a,f_b), & ||y_{a} - y_{b}||_1 \leq c_1,\\ 
\max(\cos(f_a,f_b) - m_1, 0), & ||y_{a} - y_{b}||_1 > c_2,
\end{cases}
\end{multline}
\begin{multline}\label{equation-loss-x-e}
L_{x,e}(a,b)\\ = \begin{cases}
1 - \cos(f_a,f_b), & ||e_{a} - e_{b}||_1 \leq c_3,\\ 
\max(\cos(f_a,f_b) - m_2, 0), & ||e_{a} - e_{b}||_1 > c_4.
\end{cases}
\end{multline}
%\begin{equation}\begin{split}
%&L_{x,y}(a,b)\\&=\left\{\begin{matrix}
%1 - \cos(f_a,f_b) & \text{if }  ||y_{a} - y_{b}||_1 \leq c_1\\ 
%\max(0, \cos(f_a,f_b) - m_1) & \text{if } ||y_{a} - y_{b}||_1 > c_2
%\end{matrix}\right.
%\end{split}
%\end{equation}
%\begin{equation}\begin{split} \label{equation-loss-x-e}
%&L_{x,e}(a,b)\\&=\left\{\begin{matrix}
%1 - \cos(f_a,f_b)  & \text{if } ||e_{a} - e_{b}||_1 \leq c_3\\ 
%\max(0, \cos(f_a,f_b) - m_2) & \text{if } ||e_{a} - e_{b}||_1 > c_4
%\end{matrix}\right.
%\end{split}
%\end{equation} 
The cosine similarity $\cos(f_a,f_b)=\frac{f_a\cdot f_b}{||f_a||_2||f_b||_2}$, where $\cdot$ denotes the dot product between the two vector embeddings and $||.||_p$ denotes the $\ell_p$ norm. Eqn.~(\ref{equation-loss-x-y}) defines the embedding loss based on similarity in the $Y$ space. If $y_a$ and $y_b$ are close, the cosine distance between $x_a$ and $x_b$ will be minimized. If $y_a$ and $y_b$ are far, the cosine similarity will be minimized (up to some margin $m_1 \geq 0$), thus maximizing the cosine distance. It is possible to set $c_2 > c_1$ to create a clear buffer between neighbors and non-neighbors. %We set $c_2 \geq c_1$, therefore when $c_2 > c_1$, the pairs $(a,b)$ for which $\|y_a-y_b\|_1$ is between $c_1$ and $c_2$ are not considered for the embedding. 
The loss function (\ref{equation-loss-x-e}) based on similarity in the $E$ space is exactly analogous. We combine the losses using $Y$ and $E$ similarities as
\begin{equation} \label{equation-loss-x-y-e}
L_{x,y,e}(a,b) = L_{x,y}(a,b) + w \cdot L_{x,e}(a,b),
\end{equation} where $w$ denotes the scalar weight on the $E$ loss. We set $w \leq 1$ in our experiments. The neighborhood criteria on $y$ and $e$ in (\ref{equation-loss-x-y}) and (\ref{equation-loss-x-e}) are only valid if they are continuous valued. If they are categorical, we will adopt a different neighborhood criteria, whose specifics are discussed in the relevant experiment below.

\section{Evaluation} \label{sec-eval}
To evaluate the ideas presented in this work, we focus on two fundamental questions:
\begin{enumerate}
\item Does the \ted approach provide useful explanations? \label{ques-explanation}
\item How is the prediction accuracy impacted by incorporating explanations into the training? \label{ques-accuracy}
\end{enumerate}

Since the \ted approach can be incorporated into many kinds of learning algorithms, tested against many datasets, and used in many different situations, 
a definitive answer to these questions is beyond the scope of this paper.  Instead we try to address these two questions on four datasets, evaluating accuracy in the standard way.  

Determining if any approach provides useful explanations is a challenge and no  
consensus metric has yet to emerge~\citep{finale}.  However, the \ted approach has a unique advantage in dealing with this challenge.  Specifically, since it requires explanations be provided for the target dataset (training and testing),   one can evaluate the accuracy of a model's explanation ($E$) in a similar way that one evaluates the accuracy of a predicted label ($Y$). We provide more details on the metrics used in Section~\ref{sec-eval-metrics}.  In general, we expect several metrics of explanation efficacy to emerge, including those involving the target explanation consumers \citep{tip}.

\subsection{Datasets}
The \ted approach requires a training set that contains explanations.  Since such datasets are not readily available, we evaluate the approach on a synthetic dataset (tic-tac-toe, see supplement) and leverage 3 publicly available datasets in a unique way: AADB~\citep{kong2016aesthetics}, Olfactory~\citep{olfs} and Melanoma detection~\citep{codella2018skin}.

The AADB (Aesthetics and Attributes Database)~\citep{kong2016aesthetics} contains about $10,000$ images that have been human rated for aesthetic quality ($Y \in [0, 1]$),  where higher values imply more aesthetically pleasing.  It also comes with 11 attributes ($E$) that are closely related to image aesthetic judgments by professional photographers. The attribute values are averaged over 5 humans and lie in $[-1,1]$.  The training, test, and validation partitions are provided by the authors and consist of 8,458, 1,000, and 500 images, respectively.

The Olfactory dataset~\citep{olfs}
is a challenge dataset describing various scents (chemical bondings and labels).
Each of the 476 rows represents a molecule with approximately $5000$ chemoinformatic features ($X$) 
(angles between bonds, types of atoms, etc.).  Similarly to AADB, each row also contains 21 human perceptions of the molecule, such as \emph{intensity},  \emph{pleasantness}, \emph{sour}, \emph{musky}, \emph{burnt}.  These are average values among 49 diverse individuals and lie in $[0, 100]$.
We take $Y$ to be the \emph{pleasantness} perception 
and $E$ to be the remaining 19 perceptions except for \emph{intensity}, since these 19 are known to be more fundamental semantic descriptors while pleasantness and intensity are holistic perceptions~\citep{olfs}. 
We use the standard training, test, and validation sets provided by the challenge organizers with $338$, $69$, and $69$ instances respectively.

\begin{figure}
\centering
\includegraphics[width=8.5cm]{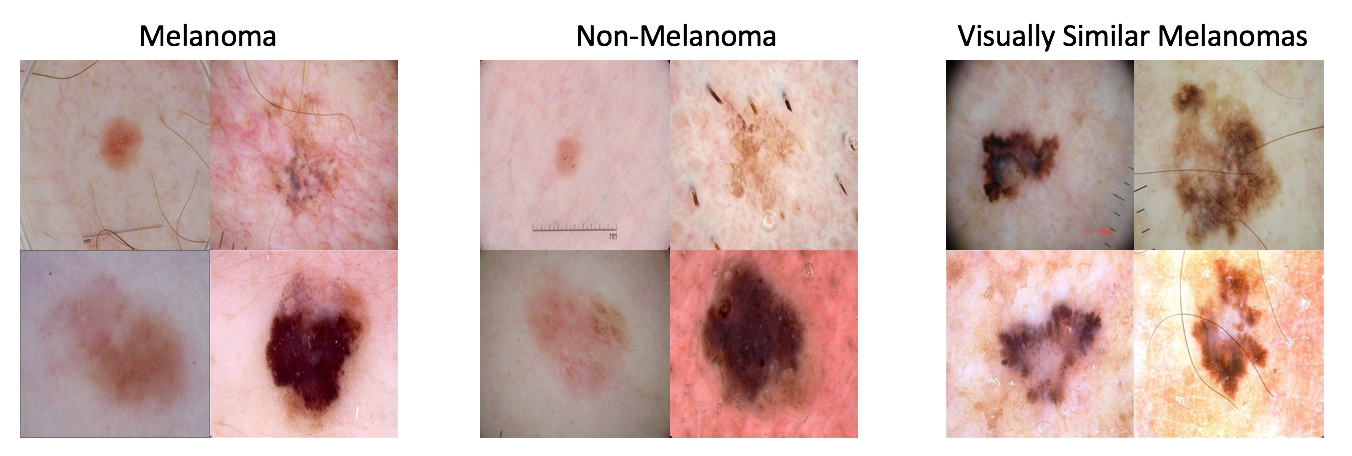}
\caption{Example images from the ISIC Melanoma detection dataset. The visual similarity between Melanoma and non-Melanoma images is seen from the left and middle images. In the right image, the visually similar lesions are placed in the same group (i.e., have the same $e$ value).}
\label{fig:melanoma_images}
\end{figure}

The 2017 International Skin Imaging Collaboration (ISIC) challenge on Skin Lesion Analysis Toward Melanoma Detection dataset \citep{codella2018skin} is a public dataset with $2000$ training and $600$ test images. Each image belongs to one of the three classes: melanoma (513 images), seborrheic keratosis (339 images) and benign nevus (1748 images). We use a version of this dataset described by \citet{codella2018collaborative}, where the melanoma images were partitioned to 20 groups, the seborrheic keratosis images were divided into 12 groups, and 15 groups were created for benign nevus, by a non-expert human user. We show some example images from this dataset in Figure \ref{fig:melanoma_images}. We take the $3$ class labels to be $Y$ and the $47$ total groups to be $E$. In this dataset, each $e$ maps to a unique $y$. We partition the original training set into a training set with $1600$ images, and a validation set with $400$ images, for use in our experiments. We continue using the original test set with $600$ images.

\subsection{Metrics}
\label{sec-eval-metrics}
An open question that we do not attempt to resolve here 
is the precise form that explanations should take. It is important that they 
match the mental model of the explanation consumer.  For example, one may expect explanations to be categorical (as in tic-tac-toe, loan approval reason codes, or our melanoma dataset) or discrete ordinal, as in human ratings. Explanations may also be continuous in crowd sourced environments, where the final rating is an (weighted) average over the human ratings. This is seen in the AADB and Olfactory datasets that we consider, where each explanation is averaged over 5 and 49 individuals respectively.

In the AADB and Olfactory datasets, since we use the existing continuous-valued attributes as explanations, we choose to treat them both as-is and discretized into $3$ bins, $\{-1, 0, 1\}$, representing negative, neutral, and positive values. The latter mimics human ratings (e.g.,~not pleasing, neutral, or pleasing).  
Specifically, we train on the original continuous $Y$ values and report absolute error (MAE) between $Y$ and a continuous-valued prediction $\hat{Y}$.  We also similarly discretize $Y$ and $\hat{Y}$ as $-1, 0, 1$.  We then report both absolute error in the discretized values (so that $\lvert 1 - 0 \rvert = 1$ and $\lvert 1 - (-1) \rvert = 2$) as well as $0$-$1$ error ($\hat{Y} = Y$ or $\hat{Y} \neq Y$), where the latter corresponds to conventional classification accuracy. 
We use bin thresholds of $0.4$ and $0.6$ for AADB and $33.66$ and $49.68$ for Olfactory to partition the $Y$ scores in the training data into thirds. 

The explanations $E$ are treated similarly to $Y$ by computing $\ell_1$ distances (sum of absolute differences over attributes) before and after discretizing to $\{-1, 0, 1\}$.  
We do not, however, compute the $0$-$1$ error for $E$.
We use thresholds of $-0.2$ and $0.2$ for AADB and $2.72$ and $6.57$ for Olfactory, which roughly partitions the values into thirds based on the training data.

For the melanoma classification dataset, since both $Y$ and $E$ are categorical, we use classification accuracy as the performance metric for both $Y$ and $E$.

\begin{table*}[ht]
\tiny
\begin{center}
\caption{Accuracy of predicting $Y$ and $E$ using different
methods (Section \ref{jointpairwise}).  Baselines for $Y$ and $E$ are regression/classification networks, Multi-task learning predicts both $Y$ and $E$ together, Embedding $Y$ + $k$NN uses the embedding from the last hidden layer of the baseline network that predicts $Y$. Pairwise $Y$ + $k$NN and Pairwise $E$ + $k$NN use the cosine embedding loss in 
(\ref{equation-loss-x-y}) and (\ref{equation-loss-x-e}) respectively to optimize the embeddings of $X$. Pairwise $Y$ \& $E$ + $k$NN uses the sum of cosine embedding losses in (\ref{equation-loss-x-y-e}) to optimize the embeddings of $X$.}
\label{table-Aes-Mel}

\subfloat[AADB dataset]{
\centering
\label{table-aesthetics}
\begin{tabular}{|c|c|c|c|c|c|c|} \hline
		& 	& \multicolumn{3}{c|}{Performance on Y} &
\multicolumn{2}{c|}{Performance on E}
    \\ \cline{3-7}
		& 	&  & \multicolumn{2}{c|}{MAE} &
\multicolumn{2}{c|}{MAE}
    \\ \cline{4-7}
Algorithm	& $\lambda$ or $k$	& Class. Accuracy	& Discretized & Continuous & Discretized &
Continuous \\ \hline \hline
Baseline ($Y$)  & NA & 0.4140 & 0.6250 & 0.1363 & NA & NA \\ \hline \hline
Baseline ($E$)  & NA
 & NA & NA & NA & 0.5053 & 0.2042 \\ \hline \hline
 & 100 & 0.4170 & 0.6300 & 0.1389 & 0.4501 & 0.1881 \\ \cline{2-7}
 & 250 & 0.4480 & 0.5910 & 0.1315 & {\bf 0.4425} & {\bf 0.1861} \\ \cline{2-7}
Multi-task  & 500 & 0.4410 & 0.5950 & 0.1318 & 0.4431 & 0.1881 \\ \cline{2-7}
regression & 1000 & {\bf 0.4730} & {\bf 0.5650} & {\bf 0.1277} & 0.4429 & 0.1903 \\ \cline{2-7}
($Y$\&$E$)  & 2500 & 0.3190 & 0.6810 & 0.1477 & 0.4917 & 0.2110 \\ \cline{2-7}
 & 5000 & 0.3180 & 0.6820 & 0.1484 & 0.5165 & 0.2119 \\ \hline \hline
 
     & 1 & 0.3990 & 0.7650 & 0.1849 & 0.6237 & 0.2724 \\ \cline{2-7}
Embedding $Y$     & 2 & {\bf 0.4020} & 0.7110 & 0.1620 & 0.5453 & 0.2402 \\ \cline{2-7}
+     & 5 & 0.3970 & 0.6610 & 0.1480 & 0.5015 & 0.2193 \\ \cline{2-7}
$k$NN     & 10 & 0.3890 & 0.6440 & 0.1395 & 0.4890 & 0.2099 \\ \cline{2-7}
     & 15 & 0.3910 & {\bf 0.6400} & 0.1375 & 0.4849 & 0.2069 \\ \cline{2-7}
     & 20 & 0.3760 & 0.6480 & {\bf 0.1372} & {\bf 0.4831} & {\bf 0.2056} \\ \hline \hline
     
	& 1 & 0.4970 & 0.5500 & 0.1275 & 0.6174 & 0.2626 \\ \cline{2-7}
Pairwise $Y$    & 2 & 0.4990 & 0.5460 & 0.1271 & 0.5410 & 0.2356 \\ \cline{2-7}
+     & 5 & 0.5040 & 0.5370 & 0.1254 & 0.4948 & 0.2154 \\ \cline{2-7}
$k$NN      & 10 & 0.5100 & 0.5310 & 0.1252 & 0.4820 & 0.2084 \\ \cline{2-7}
     & 15 & 0.5060 & 0.5320 & {\bf 0.1248} & 0.4766 & 0.2053 \\ \cline{2-7}
     & 20 & {\bf 0.5110} & {\bf 0.5290} & {\bf 0.1248} & {\bf 0.4740} & {\bf 0.2040} \\ \hline \hline
     
  & 1 & 0.3510 & 0.8180 & 0.1900 & 0.6428 & 0.2802 \\ \cline{2-7}
Pairwise $E$ & 2 & {\bf 0.3570} & 0.7550 & 0.1670 & 0.5656 & 0.2485 \\ \cline{2-7}
+ & 5 & 0.3410 & 0.7140 & 0.1546 & 0.5182 & 0.2262 \\ \cline{2-7}
$k$NN & 10 & 0.3230 & 0.6920 & 0.1494 & 0.5012 & 0.2174 \\ \cline{2-7}
 & 15 & 0.3240 & {\bf 0.6790} & 0.1489 & {\bf 0.4982} & 0.2150 \\ \cline{2-7}
 & 20 & 0.3180 & 0.6820 & {\bf 0.1483} & 0.4997 & {\bf 0.2133} \\ \hline \hline
     
% This is with weight 0.10
    & 1 & 0.5120 & 0.5590 & 0.1408 & 0.6060 & 0.2617 \\ \cline{2-7}
Pairwise $Y$ \& $E$   & 2 & 0.5060 & 0.5490 & 0.1333 & 0.5363 & 0.2364 \\ \cline{2-7}
 +    & 5 & 0.5110 & 0.5280 & 0.1272 & 0.4907 & 0.2169 \\ \cline{2-7}
$k$NN      & 10 & {\bf 0.5260} & {\bf 0.5180} & 0.1246 & 0.4784 & 0.2091 \\ \cline{2-7}
     & 15 & 0.5220 & 0.5220 & 0.1240 & 0.4760 & 0.2065 \\ \cline{2-7}
     & 20 & 0.5210 & 0.5220 & {\bf 0.1235} & {\bf 0.4731} & {\bf 0.2050} \\ \hline
\end{tabular}
}
\quad
\subfloat[ISIC Melanoma detection dataset]{
\centering
\label{table-isic-knn-accuracy}
\begin{tabular}{|c|c|c|c|c|} \hline
Algorithm	& $\lambda$ or K & Y Accuracy	& E Accuracy \\ \hline \hline
Baseline ($Y$) & NA & 0.7045 & NA   \\ \hline
Baseline ($E$) & NA & 0.6628 & 0.4107   \\ \hline \hline

 & 0.01 & 0.6711 & 0.2838  \\ \cline{2-4}
 & 0.1 & 0.6644 & 0.2838  \\ \cline{2-4}
 & 1 & 0.6544 & {\bf 0.4474}  \\ \cline{2-4}
Multi-task & 10 & 0.6778 & 0.4274  \\ \cline{2-4}
classification & 25 & {\bf 0.7145} & 0.4324  \\ \cline{2-4}
($Y$ \& $E$) & 50 & 0.6694 & 0.4057  \\ \cline{2-4}
 & 100 & 0.6761 & 0.4140  \\ \cline{2-4}
 & 250 & 0.6711 & 0.3957  \\ \cline{2-4}
 & 500 & 0.6327 & 0.3907  \\ \hline
 
 & 1 & 0.6962 & 0.2604  \\ \cline{2-4}
Embedding $Y$ & 2 & {\bf 0.6995} & 0.2604  \\ \cline{2-4}
+ & 5 & 0.6978 & 0.2604  \\ \cline{2-4}
$k$NN  & 10 & 0.6962 & 0.2604  \\ \cline{2-4}
 & 15 & 0.6978 & 0.2604  \\ \cline{2-4}
 & 20 & {\bf 0.6995} & 0.2604  \\ \hline

 & 1 & {\bf 0.6978} & 0.4357  \\ \cline{2-4}
Embedding $E$ & 2 & 0.6861 & 0.4357  \\ \cline{2-4}
+ & 5 & 0.6861 & 0.4357  \\ \cline{2-4}
$k$NN & 10 & 0.6745 & 0.4407  \\ \cline{2-4}
 & 15 & 0.6828 & {\bf 0.4374}  \\ \cline{2-4}
 & 20 & 0.6661 & 0.4424  \\ \hline 

% margin = 0.75

& 1 & 0.7162 & 0.1619  \\ \cline{2-4}
Pairwise $Y$ & 2 & {\bf 0.7179} & 0.1619  \\ \cline{2-4}
+& 5 & {\bf 0.7179} & 0.1619  \\ \cline{2-4}
$k$NN & 10 & 0.7162 & 0.1619  \\ \cline{2-4}
& 15 & 0.7162 & 0.1619  \\ \cline{2-4}
& 20 & 0.7162 & 0.1619  \\ \hline

% margin = 0.75

& 1 & 0.7245 & {\bf 0.3406}  \\ \cline{2-4}
Pairwise $E$ & 2 & 0.7279 & {\bf 0.3406}  \\ \cline{2-4}
+ & 5 & 0.7229 & 0.3389  \\ \cline{2-4}
$k$NN & 10 & 0.7279 & 0.3389  \\ \cline{2-4}
& 15 & {\bf 0.7329} & 0.3372  \\ \cline{2-4}
& 20 & 0.7312 & 0.3356  \\ \hline

\end{tabular}
}

\end{center}
\end{table*}

\begin{table*}[ht]
\tiny
\begin{center}
\caption{Accuracy of predicting $Y$ and $E$ for Olfactory using different
methods. Baseline LASSO and RF predict $Y$ from $X$. Multi-task LASSO regression with $\ell_{21}$ regularization on the coefficient matrix predicts Y\&E together, or just $E$. Other methods are similar to those in Table \ref{table-Aes-Mel}}

\label{table-olfactory}
\begin{tabular}{|c|c|c|c|c|c|c|} \hline
		& 	& \multicolumn{3}{c|}{Performance on Y} &
\multicolumn{2}{c|}{Performance on E}
    \\ \cline{3-7}
		& 	&  & \multicolumn{2}{c|}{MAE} &
\multicolumn{2}{c|}{MAE}
    \\ \cline{4-7}
Algorithm	& $k$	& Class. Accuracy & Discretized & Continuous & Discretized &
Continuous \\ \hline \hline

Baseline LASSO ($Y$)     & NA &  0.4928 & 0.5072 & {\bf 8.6483} & NA & NA  \\ \hline
Baseline RF ($Y$)       & NA &  {\bf 0.5217} & {\bf 0.4783} & 8.9447 & NA & NA  \\
\hline \hline

Multi-task regression ($Y$\&$E$) & NA & 0.4493 & 0.5507 & 11.4651 & 0.5034 &
3.6536 \\ \hline

% directly predict E from X
Multi-task regression ($E$ only) & NA &  NA & NA & NA & 0.5124 & 3.3659 \\ \hline

% Olfactory KNN

     & 1 & 0.5362 & 0.5362 & 11.7542 & 0.5690 & 4.2050 \\ \cline{2-7}
Embedding $Y$      & 2 & 0.5362 & 0.4928 & 9.9780 & 0.4950 & 3.6555 \\ \cline{2-7}
+     & 5 & {\bf 0.6087} & {\bf 0.4058} & {\bf 9.2840} & {\bf 0.4516} & {\bf 3.3488} \\ \cline{2-7}
$k$NN & 10 & 0.5652 & 0.4783 & 10.1398 & 0.4622 & 3.4128 \\ \cline{2-7}
     & 15 & 0.5362 & 0.4928 & 10.4433 & 0.4798 & 3.4012 \\ \cline{2-7}
     & 20 & 0.4783 & 0.5652 & 10.9867 & 0.4813 & 3.4746 \\ \hline \hline

 % Olfactory predict Y  (0.25)
  
  & 1 & {\bf 0.6087} & 0.4783 & 10.9306 & 0.5515 & 4.3547 \\ \cline{2-7}
Pairwise $Y$  & 2 & 0.5362 & 0.5072 & 10.9274 & 0.5095 & 3.9330 \\ \cline{2-7}
+  & 5 & 0.5507 & {\bf 0.4638} & {\bf 10.4720} & 0.4935 & 3.6824 \\ \cline{2-7}
$k$NN  & 10 & 0.5072 & 0.5072 & 10.7297 & 0.4912 & {\bf 3.5969} \\ \cline{2-7}
  & 15 & 0.5217 & 0.4928 & 10.6659 & {\bf 0.4889} & 3.6277 \\ \cline{2-7}
  & 20 & 0.4638 & 0.5507 & 10.5957 & {\bf 0.4889} & 3.6576 \\ \hline \hline

	& 1 & {\bf 0.6087} & {\bf 0.4493} & 11.4919 & 0.5728 & 4.2644 \\ \cline{2-7}
Pairwise $E$	 & 2 & 0.4928 & 0.5072 & 9.7964 & 0.5072 & 3.7131 \\ \cline{2-7}
+	 & 5 & 0.5507 & {\bf 0.4493} & {\bf 9.6680} & 0.4767 & 3.4489 \\ \cline{2-7}
$k$NN	 & 10 & 0.5507 & {\bf 0.4493} & 9.9089 & 0.4897 & 3.4294 \\ \cline{2-7}
	 & 15 & 0.4928 & 0.5072 & 10.1360 & 0.4844 & {\bf 3.4077} \\ \cline{2-7}
	 & 20 & 0.4928 & 0.5072 & 10.0589 & {\bf 0.4760} & 3.3877 \\ \hline \hline

% Ofactory predict Y Z (1.0)
 
 & 1 & {\bf 0.6522} & {\bf 0.3913} & 10.4714 & 0.5431 & 4.0833 \\ \cline{2-7}
Pairwise $Y$\&$E$ & 2 & 0.5362 & 0.4783 & {\bf 10.0081} & 0.4882 & 3.6610 \\ \cline{2-7}
+ & 5 & 0.5652 & 0.4638 & 10.0519 & {\bf 0.4622} & {\bf 3.4735} \\ \cline{2-7}
$k$NN & 10 & 0.5072 & 0.5217 & 10.3872 & 0.4653 & 3.4786 \\ \cline{2-7}
 & 15 & 0.5072 & 0.5217 & 10.7218 & 0.4737 & 3.4955 \\ \cline{2-7}
 & 20 & 0.4493 & 0.5797 & 10.8590 & 0.4790 & 3.5027 \\ \hline 

\end{tabular}
\end{center}
\end{table*}

\subsection{AADB}
We use all the approaches proposed in Section \ref{jointpairwise} to obtain results for the AADB dataset: (a) simple regression baselines for predicting $Y$ and $E$, (b) multi-task regression to predict $Y$ and $E$ together, (c) $k$NN using embeddings from the simple regression network ($Y$), (d) $k$NN using embeddings optimized for pairwise loss using $Y$ alone, and $E$ alone, and embeddings optimized using weighted pairwise loss with $Y$ and $E$.

All experiments with the AADB dataset used a modified PyTorch implementation of AlexNet for fine-tuning \citep{KrizhevskySH2012}. We simplified the fully connected layers for the regression variant of AlexNet to 1024-ReLU-Dropout-64-$n$, where $n=1$ for predicting $Y$, and $n=11$ for predicting $E$. In the multi-task case for predicting $Y$ and $E$ together, the convolutional layers were shared and two separate sets of fully connected layers with $1$ and $11$ outputs were used. The multi-task network used a weighted sum of regression losses for $Y$ and $E$: $\text{loss}_Y + \lambda \text{loss}_E$. All these single-task and multi-task networks were trained for $100$ epochs with a batch size of 64. The embedding layer that provides the $64-$dimensional output had a learning rate of $0.01$, whereas all other layers had a learning rate of $0.001$. For training the embeddings using pairwise losses, we used $100,000$ pairs chosen from the training data, and optimized the loss for $15$ epochs. The hyper-parameters $(c_1,c_2,c_3,c_4,m_1,m_2,w)$ were defined as $(0.1,0.3,0.2,0.2,0.25,0.25,0.1)$. These parameters were chosen because they provided a consistently good performance in all metrics that we report for the validation set.

Table~\ref{table-aesthetics} provides accuracy numbers for $Y$ and $E$ using the proposed approaches. Numbers in bold are the best for a metric among an algorithm. Improvement in accuracy and MAE for $Y$ over the baseline is observed for for Multi-task, Pairwise $Y$ + $k$NN and Pairwise $Y$ \& $E$ + $k$NN approaches. Clearly, optimizing embeddings based on $Y,$ and sharing information between $Y$ and $E$ is better for predicting $Y$. The higher improvement in performance using $Y$ \& $E$ similarities can be explained by the fact that $Y$ can be predicted easily using $E$ in this dataset. Using a simple regression model, this predictive accuracy was $0.7890$ with MAE of $0.2110$ and $0.0605$ for Discretized and Continuous, respectively. There is also a clear advantage in using embedding approaches compared to multi-task regression.

The accuracy of $E$ varies among the three $k$NN techniques with slight improvements by using pairwise $Y$ and then pairwise $Y$ \& $E$. Multi-task regression performs better than embedding approaches in predicting $E$ for this dataset.

\subsection{Melanoma}
For this dataset, we use the same approaches we used for the AADB dataset with a few modifications. We also perform $k$NN using embeddings from the baseline $Z$ network, and we do not obtain embeddings using weighted pairwise loss with $Y$ and $E$ because there is a one-to-one map from $E$ to $Y$ in this dataset. The networks used are also similar to the ones used for AADB except that we use cross-entropy losses. The learning rates, training epochs, and number of training pairs were also the same as AADB. The hyper-parameters $(m_1,m_2)$ were set to $(0.75, 0.75)$, and were chosen to based on the validation set performance. For the loss (\ref{equation-loss-x-y}), $a$ and $b$ were said to be neighbors if $y_a = y_b$ and non-neighbors otherwise.  For the loss (\ref{equation-loss-x-e}), $a$ and $b$ were said to be neighbors if $z_a = z_b$ and non-neighbors $y_a \neq y_b$. The pairs where $z_a \neq z_b$, but $y_a = y_b$ were not considered.

Table~\ref{table-isic-knn-accuracy} provides accuracy numbers for $Y$ and $E$ using the proposed approaches. Numbers in bold are the best for a metric among an algorithm.  The $Y$ and $E$ accuracies for multi-task and $k$NN approaches are better than that the baselines, which clearly indicates the value in sharing information between $Y$ and $E$. The best accuracy on $Y$ is obtained using the Pairwise $E$ + $k$NN approach, which is not surprising since $E$ contains $Y$ and is more granular than $Y$. Pairwise $Y$ + $k$NN approach has a poor performance on $E$ since the information in $Y$ is too coarse for predicting $E$ well.

\subsection{Olfactory}
Since random forest was the winning entry on this dataset \citep{olfs}, we used a random forest regression to pre-select $200$ out of $4869$ features for subsequent modeling. From these $200$ features, we created a base regression network using fully connected hidden layer of 64 units (embedding layer), which was then connected to an output layer. No non-linearities were employed, but the data was first transformed using $\log10(100+x)$ and then the features were standardized to zero mean and unit variance. Batch size was 338, and the network with pairwise loss was run for $750$ epochs with a learning rate of $0.0001$.  For this dataset, we set $(c_1,c_2,c_3,c_4,m_1,m_2,w)$ to $(10,20,0.0272,0.0272,0.25,0.25,1.0)$. The parameters were chosen to maximize performance on the validation set.

Table~\ref{table-olfactory} provides accuracy numbers in a similar format as Table~\ref{table-aesthetics}. The results show, once again, improved $Y$ accuracy over the baseline for Pairwise $Y$ + $k$NN and Pairwise $Y$ \& $E$ + $k$NN and corresponding improvement for MAE for $Y$. Again, this performance improvement can be explained by the fact that the predictive accuracy of $Y$ given $E$ using the both baselines were $0.8261$, with MAEs of $0.1739$ and $3.4154$ ($4.0175$ for RF) for Discretized and Continuous, respectively. Once again, the accuracy of $E$ varies among the 3 $k$NN techniques with no clear advantages. The multi-task linear regression does not perform as well as the Pairwise loss based approaches that use non-linear networks.

% Y from Z & LASSO & 0.8261 & 0.1739 & 3.4154 \\ \cline{2-5}
% 	 & RF    & 0.8261 & 0.1739 & 4.0175 \\ \hline

\section{Discussion}

One potential concern with the TED approach is the additional labor required for adding explanations.  However, researchers~\citep{zaidan-eisner:2008:gen,DBLP:journals/corr/ZhangMW16,McDonnel16why-relevant} have
% However, researchers~\citep{Zaidan07using-annotator,zaidan-eisner:2008:gen,DBLP:journals/corr/ZhangMW16,McDonnel16why-relevant} have
quantified that the
time to add labels and explanations is often the same as just adding labels for an expert SME.  They also cite other benefits of adding explanations, such
as improved quality and consistency of the resulting training data set.

Furthermore, in some instances, the $k$NN instantiation of TED may require no extra labor. For example, in cases where embeddings are used as search criteria for 
evidence-based predictions of queries, end users will, on average, naturally interact with search results that are similar to the query in explanation space. This query-result interaction activity inherently provides similar and dissimilar pairs in the explanation space that can be used to refine an embedding initially optimized for the predictions alone. This reliance on relative distances in explanation space is also what distinguishes this method from multi-task learning objectives, since absolute labels in explanation space need not be defined.

\section{Conclusion} \label{sec-conc}
The societal demand for ``meaningful information'' on 
automated decisions has sparked significant research in AI explanability.
This paper suggests a new paradigm for providing explanations from machine learning algorithms.  This new approach is particularly well-suited for explaining a machine learning prediction when all of its input features are inherently incomprehensible to humans, even to deep subject matter experts. 
The approach augments training data collection beyond features and labels to also include elicited explanations.  Through this simple idea, we are not only able to provide useful explanations that would not have otherwise been possible, but we are able to tailor the explanations to the intended user population by eliciting training explanations from members of that group.

There are many possible instantiations for this proposed paradigm of teaching explanations.  We have described a novel instantiation that learns feature embeddings using labels and explanation similarities in a joint and aligned way to permit neighbor-based explanation prediction.  We present a new objective function to learn an embedding to optimize $k$-nearest neighbor search for both prediction accuracy as well as holistic human relevancy to enforce that returned neighbors present meaningful information. We have demonstrated the proposed paradigm and two of its instantiations on a tic-tac-toe dataset (see Supplement) that we created, a publicly-available image aesthetics dataset \citep{kong2016aesthetics}, a publicly-available olfactory pleasantness dataset \citep{olfs} and a publicly-available Melanoma detection dataset \citep{codella2018skin}. We hope this work will inspire other researchers to further enrich this paradigm.

\bibliography{IEEEabrv,ted,ExAbsent}
\bibliographystyle{aaai}

% USED IN AAAI submission
%\clearpage

\appendix

\section{Synthetic Data Experiment}
We provide a synthetic data experiment using the tic-tac-toe dataset. This dataset contains the 4,520 legal non-terminal positions in this classic game.  Each position is labeled with a preferred next move ($Y$) and an explanation of the preferred move ($E$). Both $Y$ and $E$ were generated by a simple set of rules given in Section~\ref{ttt}. 

\subsection{Tic-Tac-Toe} \label{ttt}
As an illustration of the proposed approach, we describe a simple domain, tic-tac-toe, where it is possible to automatically provide labels (the preferred move in a given board position) and explanations (the reason why the preferred move is best).  A tic-tac-toe board is represented by two $3 \times 3$ binary feature planes, indicating the presence of X and O, respectively.  An additional binary feature indicates the side to move, resulting in a total of 19 binary input features.  Each legal board position is labeled with a preferred move, along with the reason the move is preferred.  The labeling is based on a simple set of rules that are executed in order (note that the rules do not guarantee optimal play):
\begin{enumerate}
\item If a winning move is available, completing three in a row for the side to move, choose that move with reason \textit{Win}
\item If a blocking move is available, preventing the opponent from completing three in a row on their next turn, choose that move with reason \textit{Block}
\item If a threatening move is available, creating two in a row with an empty third square in the row, choose that move with reason \textit{Threat}
\item Otherwise, choose an empty square, preferring center over corners over middles, with reason \textit{Empty}
\end{enumerate}

Two versions of the dataset were created, one with only the preferred move (represented as a $3 \times 3$ plane), the second with the preferred move and explanation (represented as a $3 \times 3 \times 4$ stack of planes).  A simple neural network classifier was built on each of these datasets, with one hidden layer of 200 units using ReLU and a softmax over the 9 (or 36) outputs.  
On a test set containing 10\% of the legal positions, this classifier obtained an accuracy of 96.53\% on the move-only prediction task, and 96.31\% on the move/explanation prediction task (Table~\ref{table-ttt}).  When trained on the move/explanation task, performance on predicting just the preferred move actually increases to 97.42\%.  This illustrates that the overall approach works well in a simple domain with a limited number of explanations.  Furthermore, given the simplicity of the domain, it is possible to provide explanations that are both useful and accurate.
 
\begin{table}[h]
\begin{center}
\caption{Accuracy of predicting Y, Y and E in tic-tac-toe}
\label{table-ttt}
\begin{tabular}{|c|c|c|}
\hline
Input	& Y Accuracy &  Y and E Accuracy  \\ \hline
Y & 0.9653 & NA \\ \hline
Y and E & 0.9742 & 0.9631 \\ \hline

\end{tabular}
\end{center}
\end{table}
\end{document}